\documentclass[conference]{IEEEtran}
\IEEEoverridecommandlockouts
\usepackage{cite}
\usepackage{amsmath,amssymb,amsfonts}
\usepackage{algorithmic}
\usepackage{graphicx}
\usepackage{textcomp}
\usepackage{times}
\usepackage{epsfig}
\usepackage{multirow}
\usepackage{multicol}
\usepackage{verbatim}
\usepackage{diagbox}
\def\BibTeX{{\rm B\kern-.05em{\sc i\kern-.025em b}\kern-.08em
    T\kern-.1667em\lower.7ex\hbox{E}\kern-.125emX}}
\begin{document}

\title{Detecting Adversarial Perturbations with Saliency
}

\author{\IEEEauthorblockN{Chiliang Zhang}
\IEEEauthorblockA{\textit{Institution of Microelectronics} \\
\textit{Tsinghua University}\\
Beijing, China \\
zhangcl16@mails.tsinghua.edu.cn}
\and
\IEEEauthorblockN{Zhimou Yang}
\IEEEauthorblockA{\textit{Department of Automation} \\
\textit{North Eastern University}\\
Shenyang, China \\
yzm947587207@gmail.com}

\and
\IEEEauthorblockN{Zuochang Ye}
\IEEEauthorblockA{\textit{Institution of Microelectronics} \\
\textit{Tsinghua University}\\
Beijing, China \\
zuochang@tsinghua.edu.cn}
\and
\IEEEauthorblockN{Yan Wang}
\IEEEauthorblockA{\textit{Institution of Microelectronics} \\
\textit{Tsinghua University}\\
Beijing, China \\
wangy46@tsinghua.edu.cn}
}

\maketitle

\begin{abstract}
In this paper we propose a novel method for detecting adversarial examples by training a binary classifier with both origin data and saliency data. In the case of image classification model, saliency simply explain how the model make decisions by identifying significant pixels for prediction. A model shows wrong classification output always learns wrong features and shows wrong saliency as well. Our approach shows good performance on detecting adversarial perturbations. We quantitatively evaluate generalization ability of the detector, showing that detectors trained with strong adversaries perform well on weak adversaries. 

\end{abstract}

\begin{IEEEkeywords}
Adversarial Examples, Saliency, Convolutional Neural Networks, Model Interpretation.
\end{IEEEkeywords}

\section{Introduction}
Deep Convolutional Neural Networks (CNNs) have made significant progress in classification problems \cite{krizhevsky2012imagenet,simonyan2014very,szegedy2015going,he2016deep}, which have shown to generate good results when provided sufficient data. However, CNNs are found to be easily fooled by adversarial examples generated by adding small and visually imperceptible modifications on normal samples, leading to wrong classification results.  The existence of adversarial examples reminds us rethinking differences between human visual system and computer vision system based on DNNs. Many defense methods \cite{Bastani2016Measuring, Gu2014Towards,  Huang2015Learning, Jin2015Robust, Krizhevsky2009Learning,kurakin2016adversarial, Shaham2015Understanding, Zheng2016Improving} are proposed to make neural networks more robust to adversarial examples. 

Recently, improving CNNs robustness to adversarial examples has attracted the attention of many researchers. Several defense methods are proposed to classify adversarial examples correctly, while most of these methods are easily to be attacked as well. Detection on adversarial examples is another defense task focusing on distinguish between clean samples and adversarial samples \cite{feinman2017detecting, Bhagoji2017Dimensionality, gong2017adversarial, grosse2017statistical, Metzen2017On, Dan2017Early, Li2016Adversarial}. By assuming that adversarial dataset and origin dataset are intrinsically different, classifiers are trained to determine if a sample is clean or adversarial. However, these detection are can be easily destroyed by constructing  a differentiable function that is minimized for fooling both classifier and detector with strong iterative attacks.

In this work, we adopt saliency, a tool for explaining how a classification DNN can be queried about the spatial support of a particular class in a given image, to tackle with detecting adversarial examples. To calculate saliency for an output w.r.t. input image, we use calculations with gradients to figure out importance of each individual pixels which is meant to reflect their influence on the final classification. Notice that a model learns wrong classification output always learns wrong features and wrong saliency as well. Using the DNN's intrinsic quality that adversarial samples don't completely match it's saliency guides us training a binary classifier to distinguish between real and fake samples.

\section{Background}

Formally, given an image x with ground truth $y = f_{\theta}(x)$, non-targeted adversarial example $x^*$ targeted adversarial example $x_t^*$  for target $t$ are suppose to satisfy the following constraints:
\begin{equation}
f_{\theta}(x^*) \neq  y 
\end{equation}
\begin{equation}
f_{\theta}(x_t^*) = t
\end{equation}
\begin{equation}
d(x, x^*) \leq B
\end{equation}
where function $d$ denote distance metric to quantify similarity and $B$ denote upper bound of allowed perturbation $\epsilon$ to origin image.

In the case of DNNs, the classification model $f_{\theta}$ is a highly non-linear function. To seek out which pixels leading to wrong classification when given adversarial sample,  $f_{\theta}$ is usually approximated as a linear function:
\begin{equation}
f_{\theta}(x) =\theta_w x+\theta_b
\end{equation}
The image-specific class saliency can be calculated as the derivative of $f_{\theta}$ w.r.t. input at the image x. 
\begin{equation}
\theta_w = \frac{\partial f_{\theta}(x) }{\partial x}
\end{equation}
 The computation of the image-specific saliency map for a single class is extremely quick, since it only requires a single back-propagation pass.

\subsection{Crafting Adversarial Examples}

\textbf{Fast Gradient Sign Methods (FGSM) and Iterated Fast Gradient Sign Methods}.  \cite{goodfellow2014explaining} proposed a simple gradient based algorithm to generate adversarial examples. With a hyper-parameter step-width $\varepsilon $, adversarial example can be generated by performing one step in the direction of the gradient’s sign:
\begin{equation}
x^* = x+\varepsilon \cdot  sign(\frac{\partial f_{\theta}(x)}{\partial x})
\end{equation}
FGSM is a weak attack which is not designed for generating the minimal adversarial perturbations.
\cite{kurakin2016adversarial} introduced an iterative version of the fast gradient sign methods, where replace step-width $\varepsilon $ with multiple smaller steps $\alpha$ and setting clip value $\varepsilon $ for accumulated perturbations in all iterations. Iterated FGSM start by setting $x_0^* = x$, and for each iteration $i$ computing $x_0^*$ with:
\begin{equation}
x_i^{*} = clip_{\varepsilon}(x_{i-1}^{*} + \alpha \cdot sign(\frac{\partial f_{\theta}(x)}{\partial x}) ) 
\end{equation}

\textbf{Jacobian-based Saliency Map Approach (JSMA)}.  \cite{Papernot2015The} proposed a greedy algorithm using the Jacobian to determine choosing which pixel to be perturbed.
\begin{equation}
\label{eq:1}
\begin{aligned}
s_t &= \frac{\partial t }{\partial x_i};\\
s_o &= \sum_{j \neq t}\frac{\partial j }{\partial x_i};\\
s(x_i) &= s_t\left | s_o \right | \cdot (s_t<0) \cdot (s_o>0)
\end{aligned}
\end{equation}

In Equation \ref{eq:1}, $s_t$ represents the Jacobian of target class $t$ w.r.t. input image and $s_o$ represents sum of Jacobian values of all non-target class.  Changing the selected pixel will significantly increase the likelihood of the model labeling the image as the target class. Clearly, JSMA attack works towards optimizing the $L_0$ distance metric.

\textbf{C\&W's  Attack}. \cite{carlini2017towards} proposed an attack by  approximating the solution to the following
optimization problem:
\begin{equation}
\arg \min (d(s, x+\delta) + c \cdot l(x+\delta))
\end{equation}
where $l$ is objective function for solving $f(x+\delta)=t$. In this work, we choose 
$l(x^* = \max(\max(Z(x^*)_i: i \neq t) - Z(x^*)_t, -\kappa )$, where $\kappa $ is the hyper-parameter controlling the confidence of misclassification.

\subsection{Detecting Adversarial Examples}

A ``$N+1$'' classification model $D$ is trained \cite{grosse2017statistical}  to detect adversarial examples with the method of adding adversaries to the training set, assigning a new $N+1st$ label for them.  However, experiments in \cite{carlini2017adversarial} shows that this detection failed distinguishing adversarial examples at nearly $0 \%$ accuracy under a second round attack. Experiment in in \cite{carlini2017adversarial} also shows that this detection methods cannot resist black-box attack where attackers have no access to D. By splitting training set in half for individually training two models, D and imitated D,  C\&W's  Attack succeed $98\%$ when fooling D using parameters for attacking imitated D.

Similar to approach in \cite{grosse2017statistical}, \cite{gong2017adversarial} constructs a ``$1+1$'' classification model by means of regarding real data and fake data as two completely different datasets despite being visually similar. Because of the intrinsic similarity between``$N+1$'' detection model and ``$1+1$'' detection model, this method also failed at second round attack in nearly $0\%$ accuracy for detecting adversarial examples. Black-box attack on ``$1+1$'' doesn't show  significant difference with ``$N+1$'' .

Extending approaches in \cite{grosse2017statistical} and \cite{gong2017adversarial} to features inside networks, \cite{metzen2017detecting} augment the base network by adding subnetworks as branches at some layers and produce an output $p_{adv} \in [0,1]$ representing the probability of the input being adversarial. By training the subnetworks with a balanced binary classification dataset consist of clean data and fake data generated by attacking freezed base network, the subnetwork can detect adversarial examples at the inner convolutional layers of the network. Similar to above two second round attacking methods, \cite{metzen2017detecting} propose an iterative calculating methods:
$$x_0^{adv} = x $$
\begin{equation}
\label{eq:eq10}
\begin{aligned} 
x_{n+1}^{adv} &= clip_x^{\epsilon }  \{ x_n^{adv}+\alpha[(1-\sigma ) \cdot \emph{sign} (\nabla_x J_f(x_n^{adv}\\&,y_{true}(x))) 
+ \sigma \cdot \emph{sign}(\nabla_x J_d(x_n^{adv},1))] \}
\end{aligned} 
\end{equation}
Parameter $\sigma$ is used for trading off objective for base classifier $f$ and objective for detection classifier.

\subsection{Gradients as Saliencys}

A common approach to understanding the decisions of image classification systems is to find regions of an image that were particularly influential to the final classification \cite{baehrens2010explain,zeiler2014visualizing,springenberg2014striving,zhou2016learning,selvaraju2016grad,zintgraf2016new}. At visual level, saliency represents discriminative pixels for model making decisions and \cite{simonyan2013deep} launches weakly supervised object segmentation experiment by only relying on saliency map. Saliency of wrong decision caused by fake sample always visually different from Saliency derived from right sample.

Detection method in \cite{metzen2017detecting} works well because feature maps of fake samples differ from features maps of clean samples in data distribution. We simplify this detection method by only training detector with first layer feature maps for clean data and fake data. This detector achieve 99.1\% False positive rate and 99.0\% True positive rates on detecting adversarial examples in spite of the simplification strategy. However, these two detection method cannot withstand a second-round attack which fool both original classification model and binary classifier detector. We apply attacking method in Equation \ref{eq:eq10} generating adversarial examples and lead to nearly 100\% successful attacking rate. By training detector with both feature maps in first layer and corresponding Jacobian w.r.t input image, existing second-round attack method cannot generate effective adversary directly.

\section{Methodology}
\label{gen_inst}
\begin{figure}[!h]

   \begin{center}
     \includegraphics*[width=1.0 \linewidth]{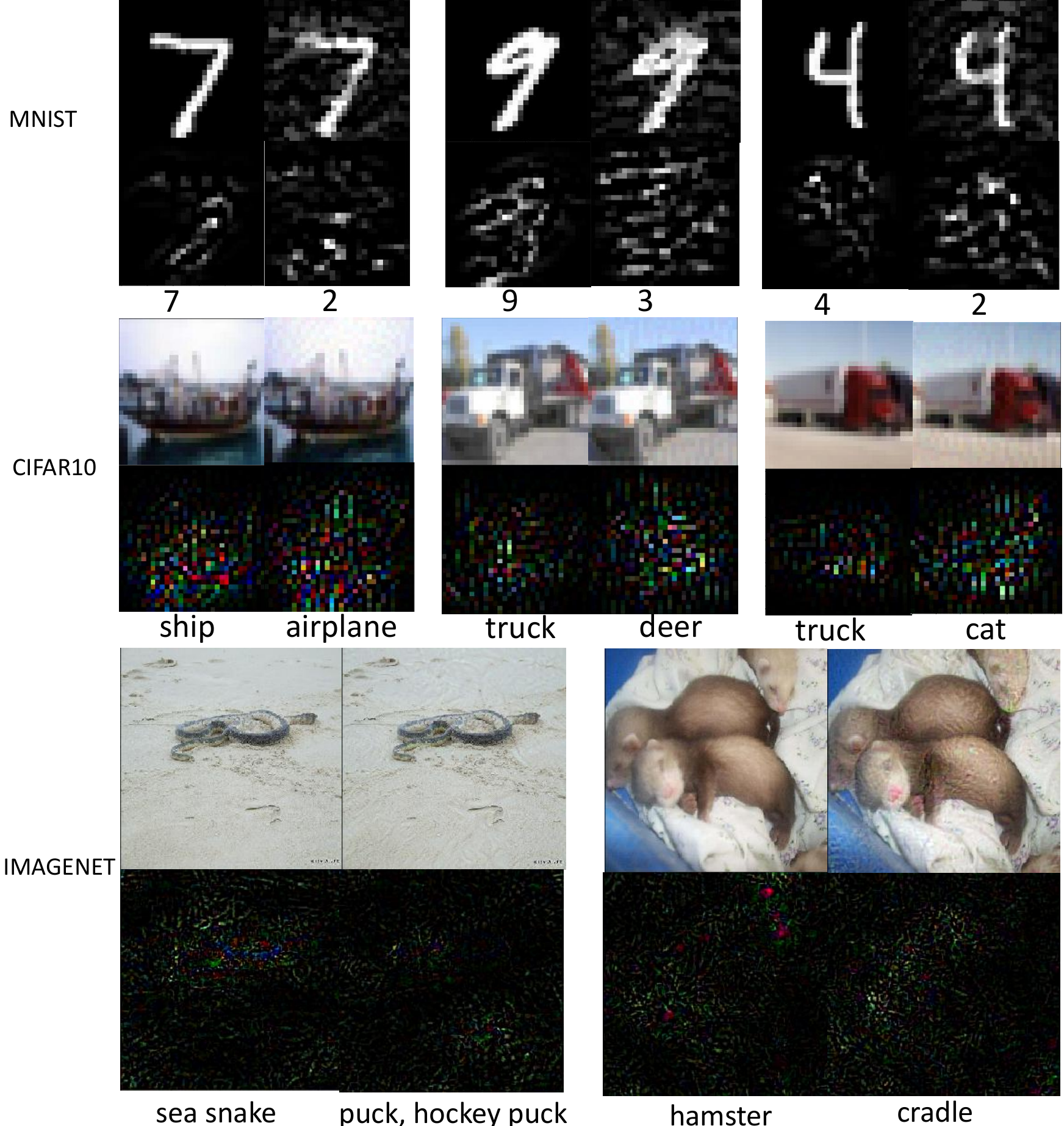}
  \caption{Origin image from MNIST,CIFAR10,Imagenet dataset and their corresponding saliency. For each four-grids sample, left parts display clean data and right parts display fake data attacked by FGSM. Lower half in four-grids sample represent corresponding saliency of upper half images.} \vspace{-0.6cm}
   \label{fig:mnist-cifar1}
   \end{center}
\end{figure}

In this section, we provide technique steps for building detectors trained with both information in pixel space and information in saliency. As is shown in Figure. \ref{fig:mnist-cifar1}, when an image is perturbed with given attacking method, saliency of classification output w.r.t. adjusted image is perturbed as well. Intuitively, attackers are supposed to keep matching for perturbed image and corresponding saliency when attacking this detector.Accordingly, We follow the steps below building our detection system.

Step1. Train a classifier $F$ with origin training dataset $X_{train}$, then craft adversarial dataset $X_{train}^{adv}$ and $X_{test}^{adv}$ by attacking $F$ using FGSM/Iterated FGSM/JSMA/C\&W.

Step2. By calculating saliency for each image in $X_{train}$, $X_{test}$, $X_{train}^{adv}$ and $X_{test}^{adv}$ based on the attacked classifier $F$, we create saliency dataset $S_{train}$, $S_{test}$, $S_{train}^{adv}$ and $S_{test}^{adv}$.

Step3. We apply both raw data and saliency data as input for training binary classifier $D$. Raw data and saliency data are concatenated on channel axis in our experiment. 

We evaluate false positive and true positive rates of detector. Furthermore, we evaluate two kinds of generalizability of $D$: 1) Attacked by the same adversary with different $\epsilon$ and 2) Attacked by one adversary when tested on data from other adversaries when fixing $\epsilon$. 

Intuitively, assuming that parameters of classifier $F$ and detector $D$ are provided, one can attack $F$ and $D$ at the same time, called Second-round Attack, which have been expressed in Equation \ref{eq:eq10}. However, traditional Second-round attack methods, eg. Equation \ref{eq:eq10} are not applicable for attacking our detection model.

\section{Experiment}
In this section, we present result of accuracy on detecting adversarial samples generated with FGSM, Iterative $l_1$, Iterative $l_{\infty}$ and C\&w attack with 3 dataset: MNIST, CIFAR10, Imagenet subset \cite{russakovsky2015imagenet}. We evaluate generalizability of $D$ for the same attack on $F$ with different choices of $\epsilon$. We also evaluate generalizability of $D$ for the same perturbation extent $\epsilon$  with different attacking methods on $F$. 

\begin{table*}[!ht]
\newcommand{\tabincell}[2]{\begin{tabular}{@{}#1@{}}#2\end{tabular}}
  \centering  
     \caption{Deep neural network used in our implementation for different datasets, called MNIST-NET-F, MNIST-NET-D, CIFAR10-NET-F, CIFAR10-NET-D, VGG16-F and VGG16-D in following passage. MNIST-NET-F, MNIST-NET-D, CIFAR10-NET-F, CIFAR10-NET-D are trained from scratch, and the left two are finetuned with VGG parameters from Caffe Model Zoo. All pooling operations and activations are set to maxpooling and relu respectively, which are not shown in this table for brevity. }
       \label{tab:table6}
  \begin{tabular}{l|c|c|c}
  \hline
    Dataset.  & MNIST & CIFAR10 10-CLASS & Imagenet\\
    \hline
    Classifier(F) & \tabincell{l}{Input(28,28,1),Conv(3,3,1,32),\\Conv(3,3,32,32),Conv(3,3,32,64),\\Conv(3,3,64,64),Fc(12544,100),Fc(100,10)}
    &\tabincell{l}{Input(32,32,3),Conv(3,3,3,32),\\Conv(3,3,32,32),Conv(3,3,32,64),\\Conv(3,3,64,64),Fc(16384,100),Fc(200,10)}&\tabincell{l}{Input(224,224,3),VGG-block1,\\VGG-block2,VGG-block3,\\Fc(25088,4096),Fc(4096,4096),Fc(4096,10)} \\
     \hline
    Detector(D) & \tabincell{l}{Input(28,28,2),Conv(3,3,2,32),\\Conv(3,3,32,32),Conv(3,3,32,64),\\Conv(3,3,64,64),Fc(12544,100),Fc(100,1)} &\tabincell{l}{Input(32,32,3),Conv(3,3,3,32),\\Conv(3,3,32,32),Conv(3,3,32,64),\\Conv(3,3,64,64),Fc(16384,100),Fc(200,1)} &\tabincell{l}{Input(224,224,6),VGG-block1,\\VGG-block2,VGG-block3,\\Fc(25088,4096),Fc(4096,4096),Fc(4096,2)}  \\
     \hline
  \end{tabular}
  \vspace{-0.2cm}
\end{table*}

\subsection{Implementation Details}
Our experiment is implemented with Keras 2.0 and Tensorflow 1.0\cite{abadi2016tensorflow}. All attacks are built with an adversarial example library cleverhans\cite{papernot2017cleverhans}.  Deep neural networks we adopt for Classifier $F$ network and Detector $D$ network are showed in Table \ref{tab:table6}. For MNIST/CIFAR10 dataset, Detector(D) network is smaller than Classifier network since intuitively binary classification for adversary detection task extract less features. Besides, all CNNs for MNIST/CIFAR10 datasets are trained from scratch. We follow \cite{metzen2017detecting} dataset collecting method, randomly selecting 10 classes from Imagenet training set and validation set, The random selected classes are: mongoose; plant, flora, plant life; Yawl; timber wolf, grey wolf, gray wolf, Canis lupus; dugong, Dugong dugon; hammer; sunglasses, dark glasses, shades; typewriter keyboard; triumphal arch; mushroom. Therefore, We have 10000 images in train set, 3035 images in validation set and 500 images(from Imagenet validation data) in test set. The motivation of using subset instead of full-dataset is of two-fold: 1) to reduce computation cost of crafting adversarial dataset, 2) to avoid adversarial conversion between similar classes, eg. perturbing image recognized as sea snake  to image recognized as water snake is not constructive. We employ VGG16 and its parameters from Caffe model zoo on initializing $F$ and $D$ for 10-CLASSES Imagenet. 

We employ 5 typical attacking algorithms in this paper: FGSM, Iterative FGSM with $l_2$ distance, Iterative FGSM with $l_{\infty}$ distance, JSMA, and C\&W attack. We revise origin FGSM to avoid label leaking problem\cite{kurakin2016adversarial}. JSMA is not applied to Imagenet subset for its' low efficiency on pixel searching when attacking images of size 224*224*3.

\subsection{MNIST/CIFAR10}

\begin{table*}[!ht]
  \centering
\caption{Accuracy on adversarial samples generated with FGSM, Iterative $l_1$, Iterative $l_{\infty}$ and C\&W attack.}
\label{sample-table}
\begin{center}
\vspace{-0.4cm}
\begin{tabular}{|c|c|c|c|c|}
 \hline 
 \multirow{2}{*}{Dataset} &
 \multicolumn{4}{c|}{FGSM/ Iterative $l_{\infty}$/ Iterative $l_2$/ C\&W attack}  \\
 \cline{2-5}
   & $f(x_{test})$ & $f(x_{test}^{adv(f)})$ & $D(x_{test})$ & $D(f(x_{test}^{adv(f)})) $
   \\
 \hline
 MNIST & 0.99/ 0.99/ 0.99/ 0.99 & 0.12/ 0.05/ 0.04/ 0.03 & 1.00/ 1.00/ 0.99/ 0.99 & 0.99/ 1.00/ 1.00/ 1.00  \\
  \hline
 CIFAR10 & 0.81/ 0.81/ 0.81/ 0.81 & 0.13/ 0.07/ 0.07/ 0.06 & 0.98/ 0.98/ 0.91 / 0.94& 0.98/ 0.98/ 0.91/ 0.95 \\
  \hline
 10-Imagenet & 0.90/ 0.90/ 0.90/ 0.81 & 0.17/ 0.09/ 0.12 /0.10& 0.92/ 0.91/ 0.93/ 0.89 & 0.90/ 0.91/ 0.94/0.84
 \\ \hline
 \end{tabular}
\end{center}
\end{table*}

\vspace{-0.2cm}

\begin{table*}[!ht]
  \centering  
     \caption{Accuracy metrics on MNIST/CIFAR10 of detector trained for adversary with maximal distortion $\epsilon_{fit}$ when tested on the same adversary with distortion $\epsilon_{test}$.}
       \label{tab:table7} \vspace{-0.2cm}
  \begin{tabular}{l|p{0.2cm}p{0.2cm}p{0.2cm}p{0.35cm}|p{0.2cm}p{0.2cm}p{0.2cm}p{0.35cm}|p{0.2cm}p{0.2cm}p{0.2cm}p{0.35cm}|p{0.2cm}p{0.2cm}p{0.2cm}p{0.35cm}|p{0.2cm}p{0.2cm}p{0.2cm}p{0.35cm}|p{0.2cm}p{0.2cm}p{0.2cm}p{0.35cm}|}
  \hline
     &\multicolumn{4}{c|}{FGSM/MNIST}&\multicolumn{4}{c|}{Iter($l_{\infty}$)/MNIST} & \multicolumn{4}{c|}{Iter($l_2$)/MNIST}&\multicolumn{4}{c|}{FGSM/CIFAR10}& \multicolumn{4}{c|}{Iter($l_{\infty}$)/CIFAR10}&\multicolumn{4}{c|}{Iter($l_2$)/CIFAR10} \\
    \hline
    \diagbox{$\epsilon_{test}$}{$\epsilon_{fit}$} & 1 &2&3&4& 1 &2&3&4& 1 &2&3&4& 1 &2&3&4& 1 &2&3&4& 1 &2&3&4\\
    \hline
    1 & 0.95 &0.88&0.71&0.65& 0.91 &0.87&0.81&0.65& 0.86 &0.80&0.76&0.70& 0.89 &0.75&0.61&0.58& 0.89 &0.82&0.77&0.65& 0.81 &0.78&0.74&0.65\\
    2 & 0.95 &1.00&0.90&0.89& 0.96 &0.96&0.96&0.92& 0.88 &0.91&0.92&0.88& 0.91 &0.95&0.86&0.81& 0.92 &0.95&0.96&0.90& 0.82 &0.89&0.82&0.78\\
    3 & 0.96 &1.00&0.99&1.00& 0.97 &0.98&0.99&0.98& 0.90 &0.93&0.95&0.95& 0.90 &0.97&0.97&0.95& 0.94 &0.98&0.98&0.99& 0.87 &0.94&0.97&0.92\\
    4 & 0.92 &0.99&0.99&1.00& 0.97 &0.99&0.99&1.00& 0.90 &0.95&0.97&1.00& 0.91 &0.92&0.99&1.00& 0.94 &0.96&0.99&0.99& 0.91 &0.94&0.98&0.99 \\
     \hline
  \end{tabular}
\end{table*}

\begin{figure*}
   \begin{center}
     \includegraphics*[width=1.0 \linewidth]{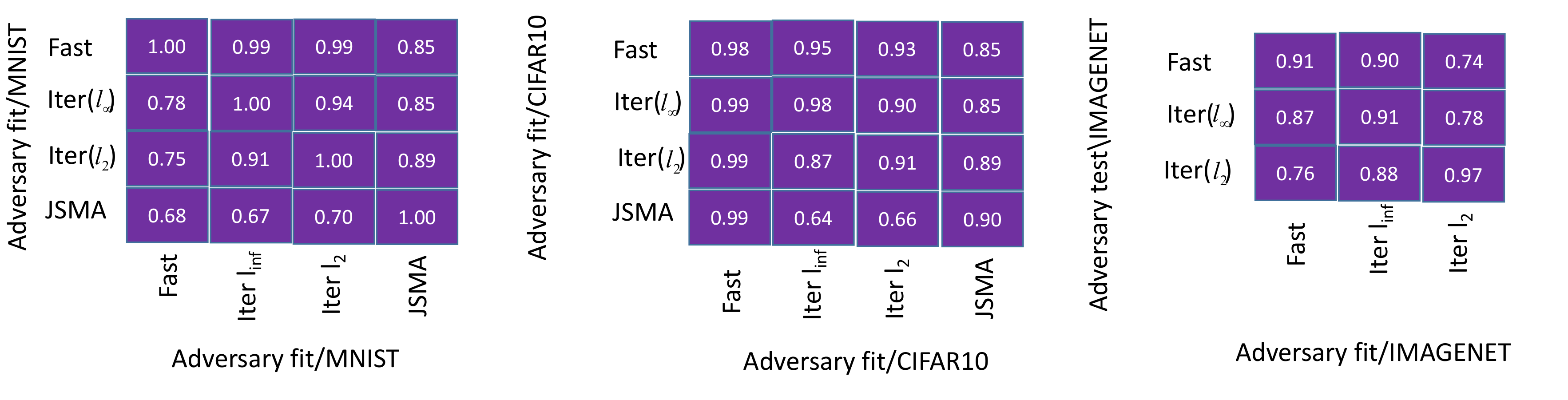}
     \vspace{-0.5cm}
  \caption{Accuracy metrics on MNIST/CIFAR10/Imagenet-subset of detector trained for one adversary when tested on other
adversaries. The maximal distortion  of the adversary (when applicable) has been chosen minimally
such that the predictive accuracy of the classifier is below $30\%$. Numbers correspond to the detection accuracy for unseen test data.} \vspace{-0.2cm}
   \label{fig:mnist-cifar2}
   \end{center}
\end{figure*}

We train MNIST-NET-F shown in Table \ref{tab:table6} for 10 epochs with Adam optimizer\cite{kingma2014adam} and learning rate was set to 0.001. MNIST-NET-F run up to 99.73\% and 99.32\% accuracy on training data and test data respectively. Afterwards, adversarial dataset was generated with 4 attacks. With clean data and adversarial data, we calculate saliency maps for all images. MNIST-NET-D are trained for 10 epochs with Adam optimizer where learning rate was set to 0.0001. CIFAR10-NET-F are trained for 100 epochs with Adam optimizer where learning rate was set to 0.0001, CIFAR10-NET-f run up to 83.89\% and 81.32\% accuracy on training data and test data respectively. CIFAR10-NET-D are trained for 5 epochs with Adam optimizer where setting learning rate as 0.0001. False positive and True positive rates of MNIST-NET-D and CIFAR10-NET-D are shown in table \ref{sample-table}.


Results in Table \ref{tab:table7} show similar performance of generalizability where a $D$ trained with large $\epsilon$ cannot reach a good effect on adversarial samples generated with small $\epsilon$. Meanwhile, $D$ trained with adversarial samples crafted with small $\epsilon$ generalized acceptably well to all adversarial samples. 

Following \cite{metzen2017detecting}, We set $\epsilon$ as minimal under the constraint that the classification accuracy is below $30\%$. Result in Figure. \ref{fig:mnist-cifar2} shows that FGSM  and JSMA  generalized not good enough with detector trained with iterative($l_2$)  and detector trained iterative($l_{\infty}$) , but iterative($l_2$) based detector and iterative($l_{\infty}$) based detector perform well to FGSM-based adversaries and JSMA-based adversaries. CIFAR10 dataset show similar character with MNIST experiment except that JSMA and FGSM cannot generalized well to each other. Therefore, we draw the conclusion for our detection approach that stronger adversary generalize well to the weak adversary since iterated method is stronger than fast method to some extent and JSMA optimize loss function under $l_0$ distance metrics where concentrated on perturbing small group of pixels severely, leading to incompatible results with other three adversary.

\subsection{Imagenet subset}
In this section, we concentrated on studying one question: if our detection approach could perform well on eye-level images. Empirically, adversarial examples on MNIST/CIFAR10 usually show visually distinguishable perturbation even texture and structure of origin image are changed. Therefore, many researches on defending MNIST/CIFAR10-level adversary helps little to find out the extrinsic difference between human visual system and deep neural networks. Take MNIST adversary for example, saliency of wrong output w.r.t. adversarial example seems visually  approximate to its' perturbation. However, in Imagenet-level images, these unreasonable properties found on MNIST/CIFAR10-level no longer appear.

\begin{table}[!ht]
  \centering  
     \caption{Accuracy metrics on Imagenet subset of detector trained for adversary with maximal distortion 
when tested on the same adversary with distortion test. Evaluation method is the same as MNIST/CIFAR10 evaluation settings.}
       \label{tab:table9}
  \begin{tabular}{l|p{0.2cm}p{0.2cm}p{0.35cm}|p{0.2cm}p{0.2cm}p{0.35cm}|p{0.2cm}p{0.2cm}p{0.35cm}|p{0.2cm}p{0.2cm}p{0.35cm}|}
  \hline
     &\multicolumn{3}{c|}{FGSM}&\multicolumn{3}{c|}{Iter($l_{\infty}$)} & \multicolumn{3}{c|}{Iter($l_2$)} \\
    \hline
    \diagbox{$\epsilon_{test}$}{$\epsilon_{fit}$} & 1 &2&3& 1 &2&3& 1 &2&3\\
    \hline
    1 & 0.84 &0.74&0.60&0.85& 0.86 &0.82&0.75&0.79& 0.79 \\
    2 & 0.89 &0.92&0.82&0.80& 0.90 &0.92&0.69&0.88& 0.88 \\
    3 & 0.89 &0.92&0.92&0.89& 0.92 &0.94&0.69&0.92& 0.92 \\
     \hline
  \end{tabular}
  \vspace{-0.3cm}
\end{table}

In this experiment, we use 3 attacking methods: FGSM, Iterative($l_1$) and Iterative($l_{\infty}$) for their suitable demand for computation recourses. We fine-tuning VGG16-F shown in Table \ref{tab:table6} for 10000 epochs with Adam optimizer\cite{kingma2014adam}.  Initial learning rate was set to 0.001, reduced to 0.0001 after 100 epochs, and further reduced to 0.00001 after 1000 epochs. VGG16-F runs up to 91.82\% and 89.83\% accuracy on training data and test data respectively. VGG16-D are trained for 100 epochs with Adam optimizer where learning rate was set to 0.0001.False positive and True positive rates of VGG16-D are shown in Table \ref{sample-table}. 

Result in Table \ref{tab:table9} shows similar direction with MNIST/CIFAR10 experiment: detectors trained with smaller perturbation upper-bound generally perform well on higher ones but not vice versa. Result in Figure. \ref{fig:mnist-cifar2} shows that detector trained with stronger adversaries generalize well to detector trained with weaker adversaries, which is identical to MNIST/CIFAR10 evaluations.

\section{Conclusion}
We have proposed a approach for detecting adversarial examples with training a binary classifier by taking saliency perturbation information into consideration. Our approach shows 100\% accuracy on detecting adversarial perturbations on MNIST dataset and show above 90\% accuracy on CIFAR10, Imagenet subset under FGSM/Iterative($l_2$)/Iterative($l_{\infty}$)/JSMA/C\&W attack. By quantitatively evaluating generalization ability of the detector, we conclude that our detector trained with strong adversaries performs well on weak adversaries, proving its' generalizability and transferability. Moreover, existing second-round attack methods cannot generate effective adversaries directly.

\bibliographystyle{IEEEtran}
\bibliography{IEEEabrv,egbib}


\end{document}